\documentclass[10pt,twocolumn,letterpaper]{article}

\usepackage{iccv}
\usepackage{times}
\usepackage{epsfig}
\usepackage{graphicx}
\usepackage{amsmath}
\usepackage{amssymb}
\usepackage{bm}
\usepackage{diagbox}
\usepackage{makecell}
\usepackage{dblfloatfix}

\usepackage{subfigure}
\usepackage{multirow}
\usepackage{array}
\usepackage{authblk}

\usepackage[pagebackref=true,breaklinks=true,letterpaper=true,colorlinks,bookmarks=false]{hyperref}

\iccvfinalcopy 

\ificcvfinal\fi

\makeatletter

\newcommand{\Rmnum}[1]{\expandafter\@slowromancap\romannumeral #1@}
\makeatother

\begin{document}

\title{GILT: Generating Images from Long Text}

\author{\textbf{Ori Bar El}} 
\author{\textbf{\;Ori Licht}} 
\author{\textbf{\;Netanel Yosephian}}

\affil{Tel-Aviv University}
\affil{\tt\small {\{oribarel, oril, yosephian\}@mail.tau.ac.il}}

\renewcommand\Authands{, }  

%

\maketitle

\begin{abstract}

Creating an image reflecting the content of a long text is a complex process that requires a sense of creativity. For example, creating a book cover or a movie poster based on their summary or a food image based on its recipe. 
In this paper we present the new task of generating images from long text that does not describe the visual content of the image directly. For this, we build a system for generating high-resolution 256 $\times$ 256 images of food conditioned on their recipes. The relation between the recipe text (without its title) to the visual content of the image is vague, and the textual structure of recipes is complex, consisting of two sections (ingredients and instructions) both containing multiple sentences. 
 We used the recipe1M~\cite{im2recipe} dataset to train and evaluate our model that is based on a the StackGAN-v2 architecture~\cite{stackgan2}.

\end{abstract}


\vspace{-8pt}
\section{Introduction}
\vspace{-5pt}

Generating images from text is a challenging task and has many applications in computer vision. Recent works have shown that Generative Adversarial Networks (GAN) are effective in synthesizing high-quality, realistic images from datasets with low
variability and low-resolution~\cite{dcgan, pyramid}. 
Further work also showed that given a text description, conditional GANs (cGAN)~\cite{conditionalgan} generate convincing images related directly to the text content~\cite{reedgan}.

\begin{figure}[h]
\begin{center}
	\includegraphics[width=0.95\linewidth]{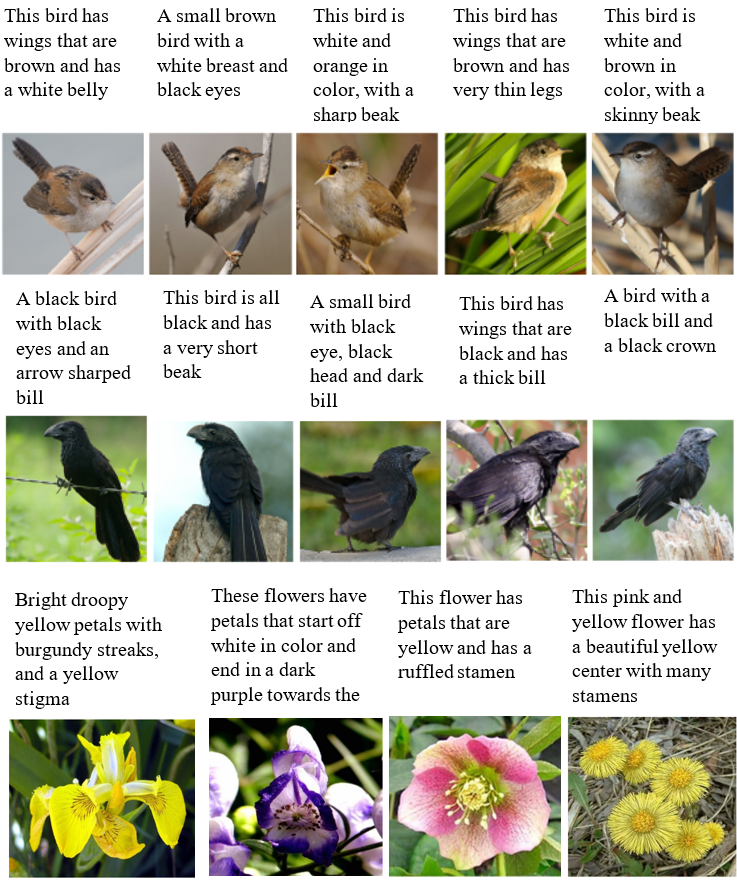}
\end{center}
\vspace{-8pt}
   \caption{Image samples from CUB and Oxford-102 datasets, and their corresponding text descriptions}
    \label{fig:big_examples}
\vspace{-8pt}
\end{figure}

\begin{figure}[b]
\begin{center}
	\includegraphics[width=1\linewidth]{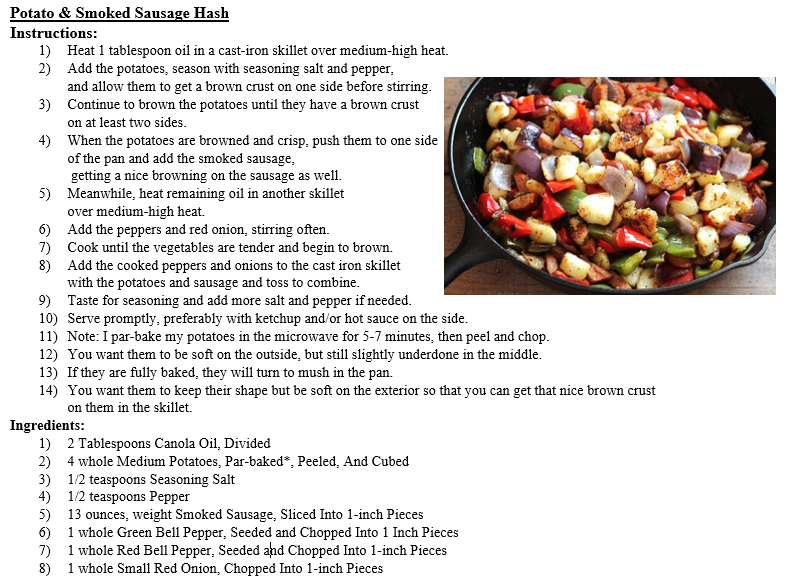}
\end{center}
\vspace{-8pt}
   \caption{Food image and its corresponding text descriptions (recipe) sampled from~\cite{im2recipe}}
    \label{fig:big_recipe}
\vspace{-8pt}
\end{figure}

All recent text to image synthesis cGANs used a short visual description of the image, with low complexity and a consistent pattern of descriptions, and the images themselves had low variability. E.g. Zhange et al.~\cite{stackgan2,stackgan1} used the CUB dataset~\cite{birdsdatatset} containing 200 bird species with 11,788 images and corresponding description and Oxford-102 dataset~\cite{flowersdataset} containing 8,189 images of flowers from 102 different categories (see Figure~\ref{fig:big_examples}).
 Recently the dataset recipe1M, containing 800K pairs of recipes and their corresponding images, was published as part of ~\cite{im2recipe}.
In comparison to the CUB and Oxford-102 datasets, this dataset has a high variability due to the variety of food categories and subcategories. Moreover, the text related to the image is complex. It consists of 2 sections (ingredients and instructions), that together might contain tens of lines (e.g. Figure~\ref{fig:big_recipe}).

We propose a novel task of synthesizing images from long text, that is related to the image but does not contain visual description of it. Specifically, We propose a baseline to this task by combining the state-of-the-art Stacked Generative Adversarial Network~\cite{stackgan2} and the two proposals of recipe embeddings computed in im2recipe~\cite{im2recipe} to generate food images conditioned on their recipes.
We also present extensive qualitative and quantitative experiments using human ranking, MS-SSIM~\cite{ms-ssim} and inception scores~\cite{inceptionscore}, to compare the effectiveness of the two embedding methods.

Our code is available at

{\href{https://github.com/netanelyo/Recipe2ImageGAN}{https://github.com/netanelyo/Recipe2ImageGAN.}}
%

\vspace{-2pt}
\section{Related Work}
 \vspace{-5pt}

Generating high-resolution images conditioned on text descriptions is a fundamental problem in computer vision. This problem is being studied extensively and various approaches were suggested to tackle it. 
Deep generative models, such as~\cite{stackgan1,stackgan2,attngan,hdgan}, achieved tremendous progress in this domain.
In order to get high-resolution images they used multi-stage GANs, where each stage corrects defects and adds details w.r.t. the previous stage.
In~\cite{attngan}, Xu et al. use Deep Attentional Multimodal Similarity Model (DAMSM) for text embedding. DAMSM learns two neural networks that map sub regions of the image and words of the sentence to a common semantic space. Thus, measures the image-text similarity at the word level to compute a fine-grained loss for image generation.

\vspace{-2pt}
\section{Learning Embeddings}
\vspace{-5pt}

Our cGAN uses the embedding of the entire recipe (except for its title) as a condition. To generate that embedding we leverage the methods used in~\cite{im2recipe}. They \cite{im2recipe} proposed two types of embedding methods, where the second adds a semantic regularization loss component. Throughout this paper we will refer to the first method without semantic regularization as NOREG, and the second with semantic regularization as REG.  The embedding methods are composed of the following steps (for the concrete architecture see the original paper).
\begin{enumerate}
    \item Preliminary embedding of the ingredients.
    \item Preliminary embedding of the cooking instructions.
    \item Joint neural embedding of the entire recipe (using the concatenation of the former preliminary embeddings) and the image into a common space, using cosine similarity loss between the embeddings of recipe-image pairs.
    \item Adding a semantic regularization loss using a high-level classification objective (used in REG only)
\end{enumerate}

We employ these methods as explained in the original paper.

\vspace{-2pt}
\section{Stacked Generative Adversarial Networks}
\vspace{-5pt}
Originally, GANs~\cite{gan} are a combination of two models that are trained to compete with each other. In the training process both the generator \(G\) and the discriminator \(D\) are trained.
\textbf{\(G\)} is optimized to reproduce images similar to the original data distribution, by generating images that are difficult for the discriminator \(D\) to differ from the true images. \textbf{\(D\)} is trained to distinguish between real images and fake synthetic ones, generated by \(G\).
This training is similar to solving a minimax of 2 players game, with the objective function,~\cite{gan}
\begin{equation}\label{eq:ganobjective}
\begin{aligned}
\min_{G} \max_{D} V(D,G) = \; & \mathbb{E}_{x \sim {p_{data}}} [\log D(x)] \; + \\
& \mathbb{E}_{z \sim {p_{z}}} [\log(1 - D(G(z)))],
\end{aligned}
\end{equation}
where $x$ is an image sampled from the real distribution $p_{data}$ and $z$ is the noise vector, which is sampled from a prior distribution \(p_z\) (e.g Uniform or Gaussian), used by $G$ to generate the synthetic image.

In the case of Conditional GANs~\cite{attngan,stackgan2,stackgan1,reedgan} both the generator and the discriminator are compelled to consider another variable $c$.
We denote $D(x,c)$ and $G(z,c)$ to be the generator $G$ and the discriminator $D$ conditioned by $c$, respectively. Meaning that $G$ is able to generate images, and $D$ discriminate them, conditioned on $c$.

The StackGAN-v2 model, introduced in StackGAN++, by Zhang et al.~\cite{stackgan2}, is an end-to-end network for modeling a series of multi-scale image distributions. The architecture of this model is consisted of several generators and discriminators in a tree-like structure framework (for the concrete architecture see the original paper).
Given a noise vector $z \sim {p_{z}}$ and condition $c$  StackGAN-v2 generates images from low-resolution to high-resolution from different branches of the tree. 
  In our case $c$ is one of the recipe embeddings from section 4. Overall, we have one model for each of the two embeddings.

\begin{figure*}[t]
\centering
\includegraphics[width=1.0\linewidth]{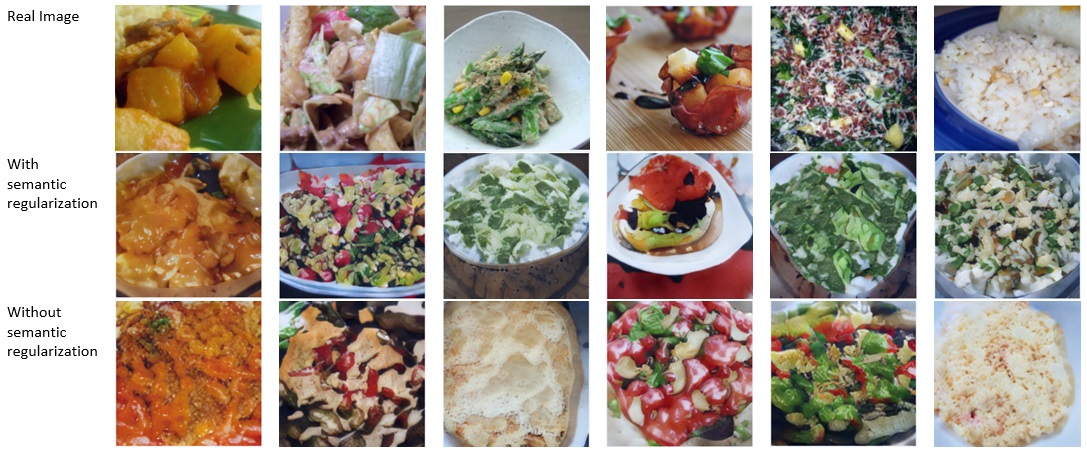}
   \caption{Comparison of the real image, the generated image using the semantic-regularization (REG), and without the regularization (NOREG), \textbf{where most humans preferred the regularized images}.}
	\vspace{-12pt}
\label{fig:cmp_sem_and_emb1}
\end{figure*}

\begin{figure*}[tb]
\centering
\includegraphics[width=1.0\linewidth]{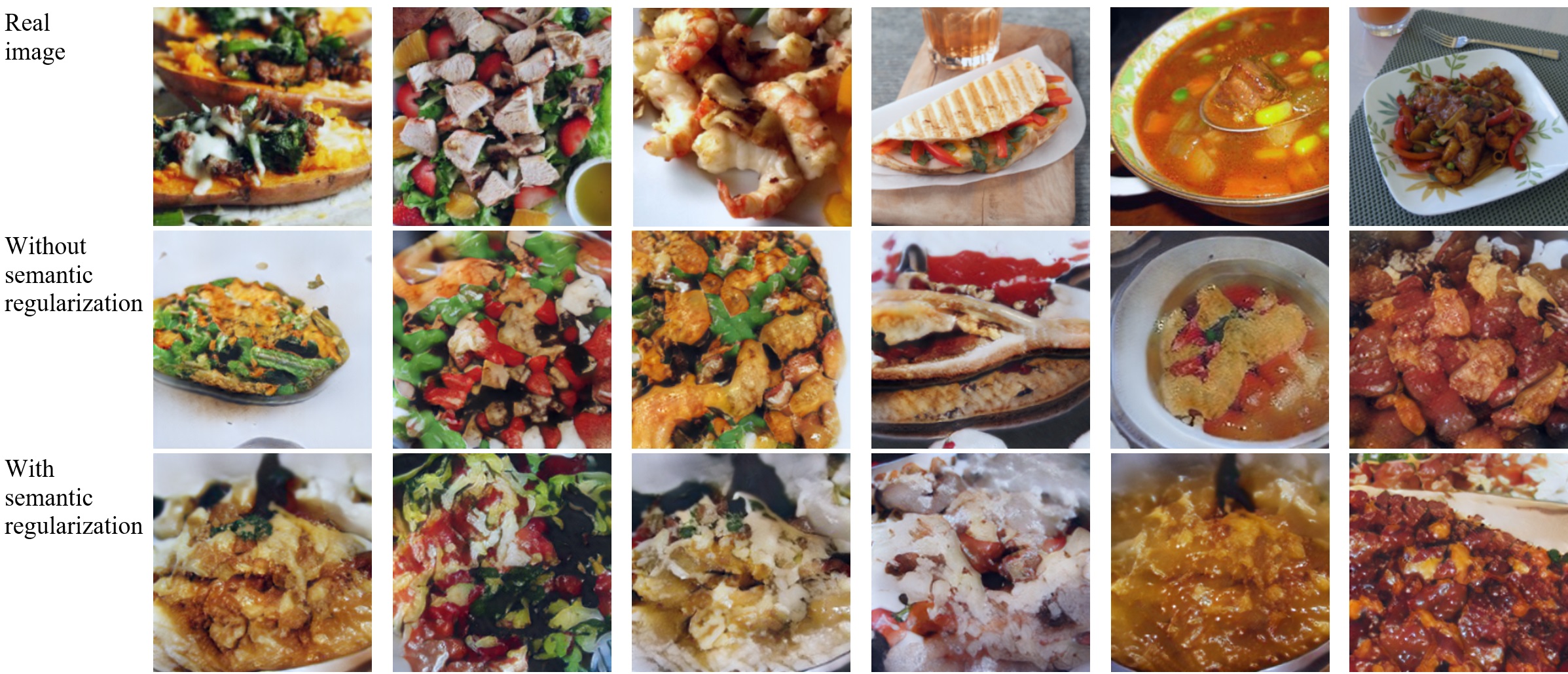}
   \caption{Comparison of the real image, the generated image without using the semantic-regularization (NOREG), and with semantic-regularization (REG), \textbf{where most humans preferred the non-regularized images}.}
	\vspace{-12pt}
\label{fig:cmp_sem_and_emb2}
\end{figure*}

\vspace{-12pt}
\begin{figure*}[tb]
\centering
\includegraphics[width=1.0\linewidth]{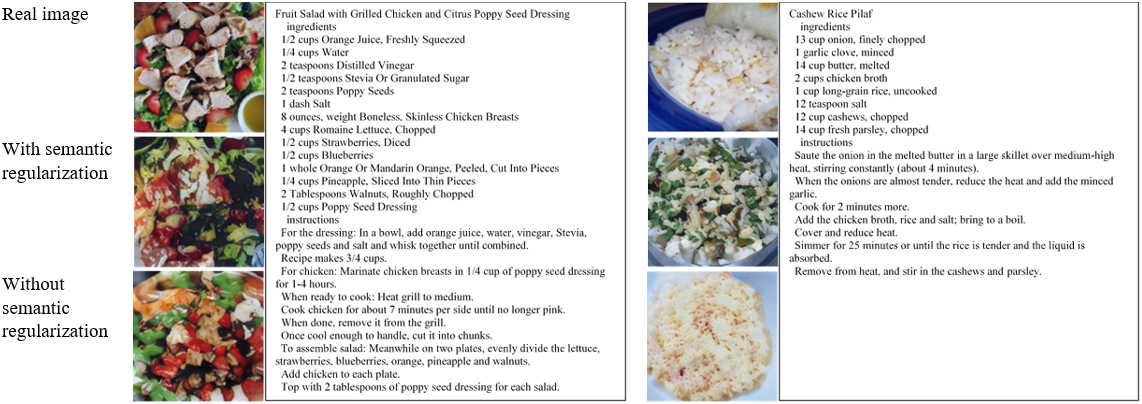}
   \caption{Comparison of the real image, the generated image with using the semantic-regularization (REG) and without semantic-regularization (NOREG), top-to-bottom, and the corresponding recipe. \bf {One can notice that in the generated image using semantic-regularization (on the right side), it has a dominant green color, which is, probably, due to the parsley that in the ingredients, and is different than the real image}.}
	\vspace{-12pt}
\label{fig:cmp_real_sem_emb}
\end{figure*}

\vspace{9pt}
\section{Implementation details}
\vspace{-5pt}
We compare two types of embedding methods from~\cite{im2recipe}. The first method is based on cosine-similarity loss only and is of size 1024. The second method uses additionally a high-level classification objective to compute embedding of size 1048. For training the StackGAN-v2~\cite{stackgan2} model we used a batch size of 24. Trying to use a larger batch size resulted in a mode-collapse.
The text-embedding dimension parameter presented in StackGAN-v2 was of size 128. At first we used this parameter with our training and got poor results. We realized that by projecting the rich text on this small dimension, we might omit discriminative subtleties between different recipes.
As a result we used 1024 as the text-embedding dimension parameter for both of the embeddings methods. In order to accelerate the training process we used hdf5 (hierarchical data format) to map files to memory.
All neural models were implemented using PyTorch framework. All other parameters are identical to~\cite{im2recipe} and~\cite{stackgan2}. The models were trained on 3 Nvidia Titan-X GPUs, each has 12GB of memory, for 100 epochs on each of the embedding methods.
\section{Experiments}
\vspace{-5pt}

To evaluate our models, we conduct quantitative, in the form of Inception Score (IS) \cite{inceptionscore}, and qualitative, in the form of Human Rankings (HR), evaluations. We compare the aforementioned evaluation methods on images generated by two different text embedding methods, which both are computed using \cite{im2recipe}. In addition, we show outputs from several state-of-the-art and previous state-of-the-art text-to-image synthesis models, which indicate that generating realistic food-images based on their description is a challenging task, and all the more so based on their recipe. Moreover, we examine the diversity of the generated images using MS-SSIM~\cite{ms-ssim}.

\vspace{-2pt}
\subsection{Datasets and evaluation metrics} 
\vspace{-5pt}

Recipe1M~\cite{im2recipe} contains over 1 million recipes and 800k food images. Due to hardware limitations, we used a training set of 52k and an evaluation set of 24k recipe-image pairs. In the pre-processing stage, the images were down-scaled from $256\times256$ to $128\times128$ and $64\times64$, in-order to train on different image scales. Further more, the images were cropped and horizontally flipped randomly. This was a best-effort to focus on the food object in the image, but from time to time, the cropping eliminated important details from the original image.

\textbf{Evaluation metrics. }
Even though evaluating generative models is often a difficult task (as mentioned in \cite{noteonIS}), to compare between the generated images in both embedding methods quantitatively (numerically) we use Inspection-Score,
\begin{equation}\label{eq:incpetionScoreEq}
\begin{aligned}
IS = \exp(\mathbb{E}_{\bm{x}} D_{KL} (p(y|\bm{x}) \,||\, p(y))),
\end{aligned}
\end{equation}
Where $\bm{x}$ denotes a single generated sample, $y$ is the predicted label, $p(y|\bm{x})$ and $p(y)$ are the conditional and marginal class distributions, respectively, and $D_{KL}$ is the Kullback-Leibler (KL) divergence. Intuitively, the IS measures the diversity, with respect to ImageNet~\cite{imagenet} classes, and clarity of the generated images. Therefore, the KL-divergence (hence the IS), of a successful generator should be large. We evaluate the IS on the evaluation set, which contains 24k randomly chosen samples. In spite of the suboptimality of IS, stated in \cite{noteonIS}, it is the most popular method to evaluate generative models.

Due to the aforementioned suboptimality of IS and the fact that it does not reflect the correlation between the generated image and the recipe it is conditioned on, a qualitative evaluation metric was used. Therefore, 30 people were asked to rank, a total of 10 samples, in several aspects:
\begin{enumerate}
    \item \label{HR-q1} The strength of the relation between a generated image and its corresponding recipe.
    \item \label{HR-q2} The strength of the relation between a generated image and its corresponding real image.
    \item \label{HR-q3} In which degree the image appears to be a real food image.
\end{enumerate}
The final human rankings are the average of the above.

\begin{table}[bt]
\begin{center}
\scriptsize
\begin{tabular}{|c|l|c|c|c|}
\hline
    \backslashbox{Metric}{Embedding\\Type} &\makecell{}
    &\makecell{REG} &\makecell{NOREG}\\
\hline
    \makecell{Inception\\Score} &\makecell{} &4.42~$\pm$~0.17 &{\bf 4.55~$\pm$~0.20} \\
\hline
    \makecell{Human\\Ranking} &\makecell{Q~\ref{HR-q1}\\Q~\ref{HR-q2}\\Q~\ref{HR-q3}} &\makecell{2.62\\2.24\\3.05} &{\bf \makecell{2.88\\2.70\\3.72}} \\
\hline
\end{tabular}
\end{center}
\vspace{-5pt}
    \caption{Inception scores and average human rankings of our results.}
\vspace{-9pt}
\label{tab:cmp_scores}
\end{table}
\begin{figure}[tb]
\center
\includegraphics[width=0.85\linewidth]{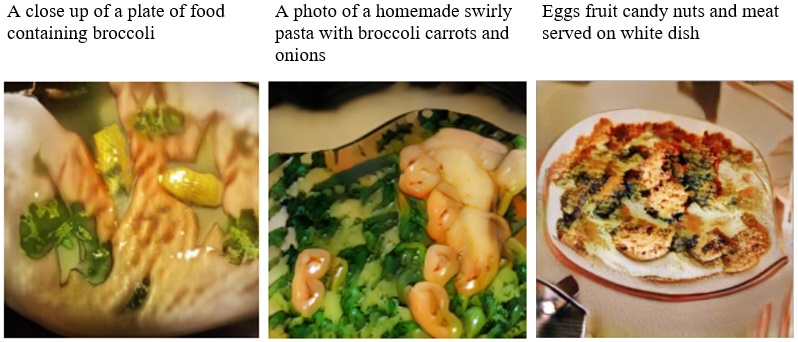}
\vspace{-2pt}
    \caption{Example results of food images generated by HDGAN~\cite{hdgan}, AttnGAN~\cite{attngan} and StackGAN++~\cite{stackgan2} conditioned on text descriptions.}
\vspace{-8pt}
\label{fig:other_papers}
\end{figure}

\vspace{-2pt}
\subsection{Quantitative and qualitative results}
\vspace{-5pt}

As one can see by the examples shown in Figure~\ref{fig:other_papers}, the state-of-the-art and previous state-of-the-art text-to-image synthesis models yields unsatisfying results, in spite of the concise and visually descriptive text on which it is conditioned.

We compare our model between the two mentioned text embedding methods, \ie, embedding with semantic regularization and without it. The inception scores and average human ranks for our models are reported in Table~\ref{tab:cmp_scores}. As can be seen in the table, the embedding without semantic regularization achieves the better IS and HR (in all aspects) scores.
Representative examples are compared in Figures~\ref{fig:cmp_sem_and_emb1}, \ref{fig:cmp_sem_and_emb2} and \ref{fig:cmp_real_sem_emb}.

\textbf{Human rankings. }
10 corresponding pairs of generated images were chosen, from each embedding method evaluation results (\ie, images that were generated conditioned on the same recipe). Our subjects were asked to rank the images in the aforementioned aspects, on a scale of 1 to 5. As mentioned earlier, the model trained conditioned on the cosine-similarity based embedding method yielded results that are close to real-like images. It is worth mentioning that there were real food images that were given less-than-or-equal rank in comparison to generated images (in the cosine-similarity embedding evaluation).

\vspace{-2pt}
\subsection{Diversity}
\vspace{-5pt}

The most successful method to evaluate image similarity, as mentioned in \cite{classifiergan}, is  multi-scale
structural similarity (MS-SSIM~\cite{ms-ssim}). The method attempts to discount the image aspects that are not important to the human eye. To evaluate the diversity of the images generated by our model, 200 random images from the evaluation set were chosen, and we calculated the MS-SSIM score for each pair. The results can be seen in Table~\ref{tab:ssim_table}. One can see that the embedding method without semantic regularization, achieved better score (lower is better), \ie, generated more diversed images. These results might be explained by the fact that - when using semantic regularization, the classification based regularization aims to map the recipe embedding to one of the 1048 classes in a discrete manner, instead of utilizing the entire space.

\begin{table}[bt]
\begin{center}
\scriptsize
\begin{tabular}{|c|l|c|c|c|}
\hline
    \makecell{Embedding Type} &\makecell{MS-SSIM score}\\
\hline
    \makecell{REG} &\makecell{0.17}\\
\hline
    \makecell{NOREG} &{\bf \makecell{0.07}}\\
\hline
\end{tabular}
\end{center}
\vspace{-5pt}
    \caption{MS-SSIM\cite{ms-ssim} score of randomly chosen images from our results.}
\vspace{-15pt}
\label{tab:ssim_table}
\end{table}

\vspace{-5pt}
\section{Conclusions}
\vspace{-5pt}

In this paper, we propose an end-to-end system  for high-resolution long text to image synthesis using Stacked Generative Adversarial Network (StackGAN-v2).
We compare two embedding types, one based on cosine-similarity (NOREG) and the second combines a high-level classification objective (REG). The proposed methods prove their ability to generate photo-realistic images of food from their text recipe (ingredients and instructions only). Herein, we provide a baseline for this novel task.
It is worth mentioning that the quality of the images in the recipe1M dataset~\cite{im2recipe} is low in comparison to the images in CUB ~\cite{birdsdatatset} and Oxford-102 datasets~\cite{flowersdataset}. This is reflected by lots of blurred images with bad lighting conditions, "porridge-like images" and the fact that the images are not square shaped (which makes it difficult to train the models). This fact might give an explanation to the fact that both models succeeded in generating "porridge-like" food images (e.g. pasta, rice, soups, salad) but struggles to generate food images that have a distinctive shape (e.g. hamburger, chicken, drinks).
From the results, it is evident that the method NOREG outperforms the method REG, by generating more vivid images with more photo-realistic details. Moreover, the inception score and diversity measures of the former is better than the latter. 
Overall, we show that although REG outperforms NOREG in a classification task (see~\cite{im2recipe}) it is inferior for generating new images.

{\small
\bibliographystyle{ieee}
\bibliography{AAAcurrent_paper}
}

\clearpage
\onecolumn
\centerline{\Large \bf Supplementary Materials}
\section*{More Results of Generated Images by Both Embedding Methods and their Corresponding Recipes.}

\includegraphics[width=0.95\linewidth]{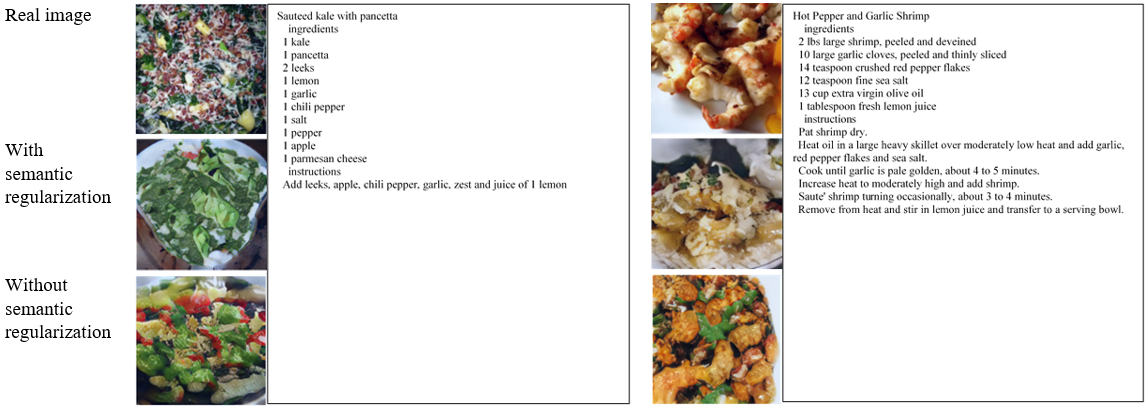}
\vspace{+10pt}

\includegraphics[width=0.95\linewidth]{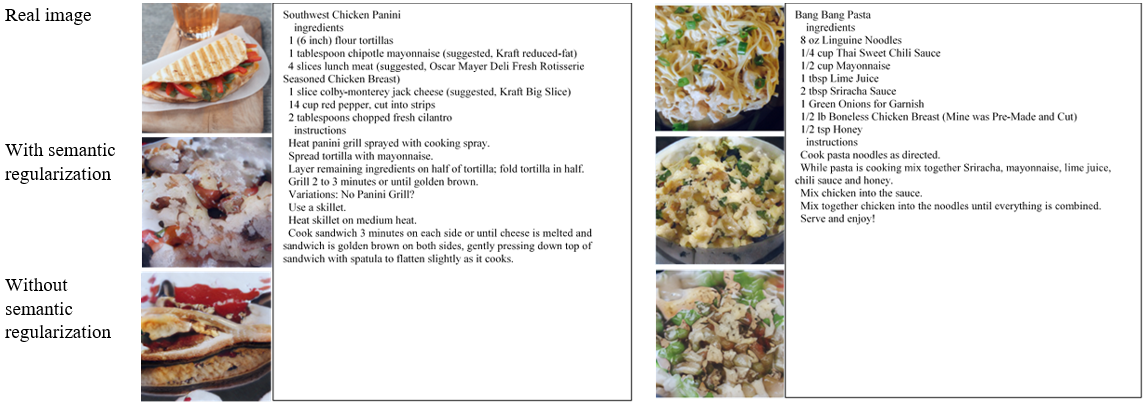}
\vspace{+10pt}

\includegraphics[width=0.95\linewidth]{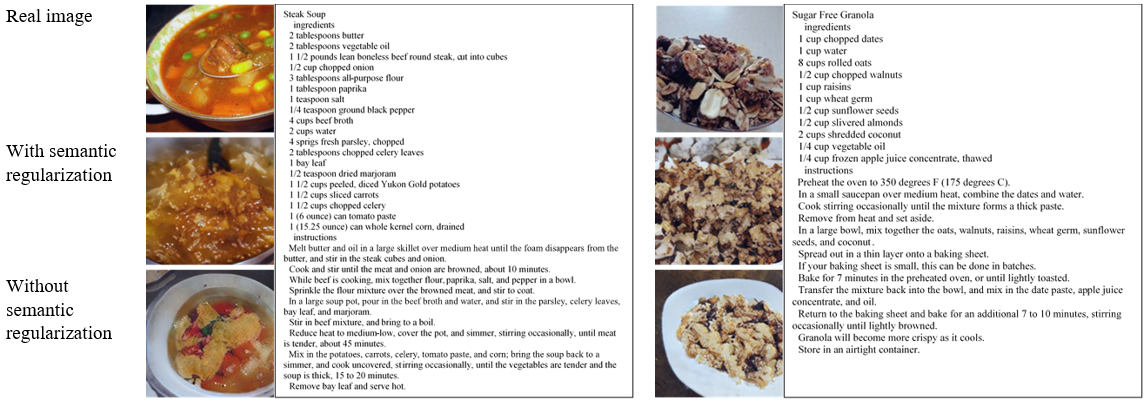}
\vspace{+10pt}

\includegraphics[width=0.95\linewidth]{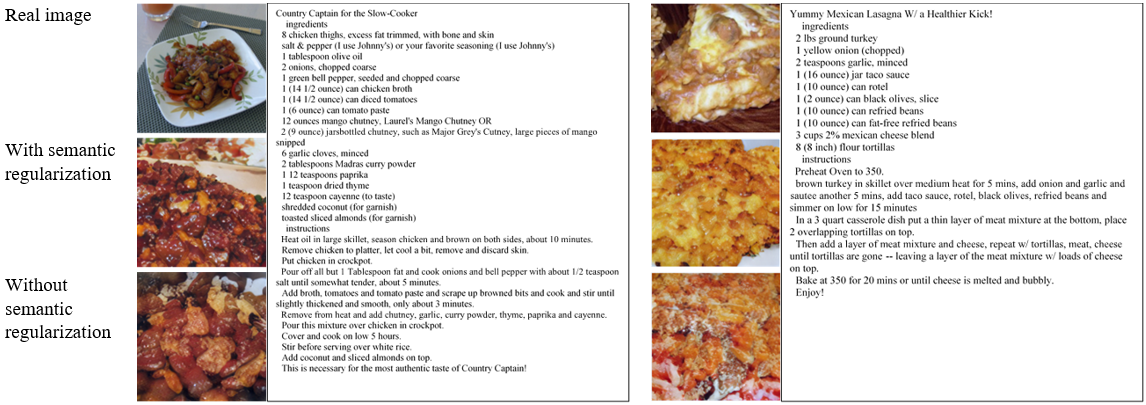}
\vspace{+10pt}

\includegraphics[width=0.95\linewidth]{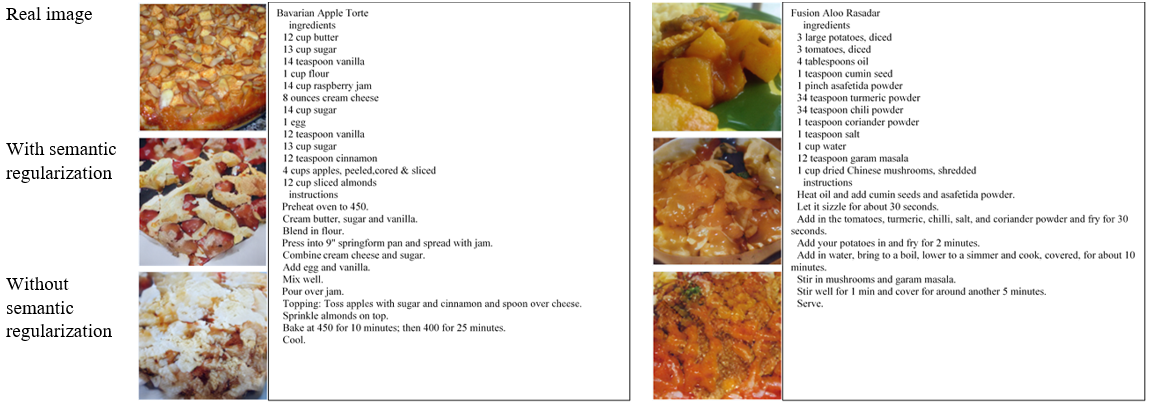}
\vspace{+10pt}

\includegraphics[width=0.95\linewidth]{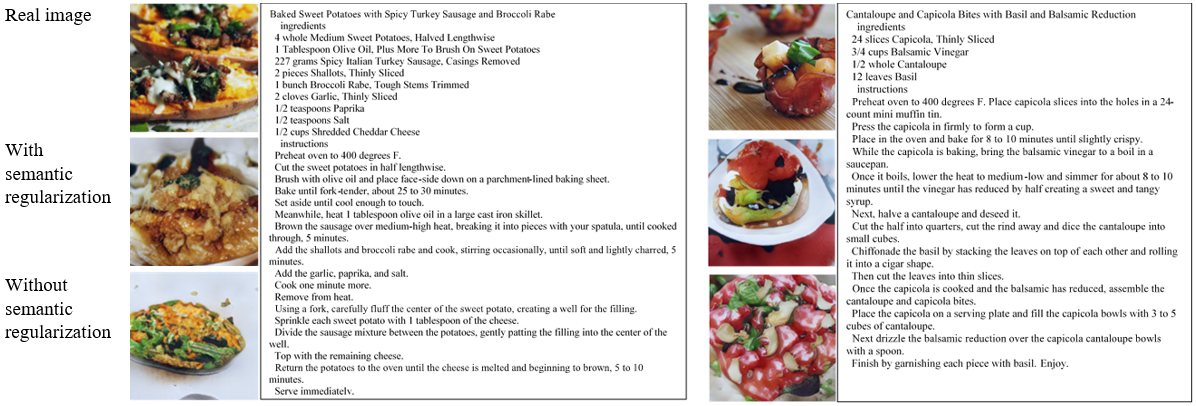}

\end{document}